\documentclass[hidelinks]{article}
\pdfoutput=1

\usepackage[preprint]{neurips_2023}

\usepackage{booktabs, graphicx, subcaption, pgfplots}
\usepackage[inline]{enumitem}
\usepackage[utf8]{inputenc} 
\usepackage[T1]{fontenc}    
\usepackage{hyperref}       
\usepackage{url}            
\usepackage{booktabs}       
\usepackage{amsfonts}       
\usepackage{nicefrac}       
\usepackage{microtype}      
\usepackage{xcolor}         

\usepackage{natbib}

\title{Can pruning make Large Language Models more efficient?}

\author{
    Sia Gholami \\
    The Institute of Electrical and Electronics Engineers, Member IEEE \\
    \texttt{gholami@ieee.org} \\
    \And 
    Marwan Omar \\
    Illinois Institute of Technology \\
    \texttt{momar3@iit.edu} \\
}

\begin{document}

\maketitle

\begin{abstract}
Transformer models have revolutionized natural language processing with their unparalleled ability to grasp complex contextual relationships. However, the vast number of parameters in these models has raised concerns regarding computational efficiency, environmental impact, and deployability on resource-limited platforms. To address these challenges, this paper investigates the application of weight pruning—a strategic reduction of model parameters based on their significance—as an optimization strategy for Transformer architectures. Through extensive experimentation, we explore various pruning methodologies, highlighting their impact on model performance, size, and computational demands. Our findings suggest that with judicious selection of pruning hyperparameters, significant reductions in model size are attainable without considerable compromise on performance. Moreover, when coupled with post-pruning fine-tuning strategies, some pruned models even exhibit enhanced generalization capabilities. This work seeks to bridge the gap between model efficiency and performance, paving the way for more scalable and environmentally responsible deep learning applications.
\end{abstract}

\section{Introduction}

In the burgeoning domain of deep learning, Transformer models~\citep{vaswani2017attention} have indisputably reshaped the landscape of natural language processing tasks. Their unprecedented capabilities in capturing intricate contextual information and establishing novel benchmarks across a myriad of applications have made them a vanguard in modern artificial intelligence~\citep{devlin2018bert}. However, with great power comes inherent complexity. These models, characterized by their massive parameter counts, have posed significant challenges for real-time applications, deployment on resource-constrained devices, and even exacerbated the environmental concerns of large-scale computations~\citep{strubell2019energy}.

Given the pressing demand for efficient and scalable deep learning solutions, model optimization techniques have emerged as pivotal endeavors in contemporary research. Among these, weight pruning stands out for its intuitive appeal and empirical success~\citep{han2015deep}. By strategically eliminating certain weights or connections based on their significance, pruning endeavors to retain, or even enhance, model performance while substantially reducing computational and memory footprints~\citep{zhu2017prune}.

Yet, the journey of weight pruning in the vast sea of Transformer architectures is still nascent, with its potentials and pitfalls only beginning to be explored. This paper delves into the intricate interplay between weight pruning and Transformer models, investigating the nuances of different pruning strategies, their ramifications on model performance, and charting the way forward for more efficient and environmentally responsible neural network architectures~\citep{gale2019state}. Through systematic analyses and comprehensive experiments, we aim to bridge the chasm between the desideratum of model efficiency and the inexorable demand for superior performance.

The implementation of weight pruning in Transformer models can have both positive and negative impacts on their performance, and these effects often depend on various factors, such as the degree of pruning, the specific pruning strategy used, and the complexity of the task at hand.

\begin{enumerate}
    \item Computational Efficiency: One of the primary benefits of weight pruning is the reduction in the number of parameters in the model, which can lead to increased computational efficiency. Fewer weights mean fewer operations during the forward and backward passes, resulting in less computation time. This advantage is particularly important when deploying models on resource-constrained devices or when processing large datasets.
    
    \item Model Size: Weight pruning can significantly reduce the model's size, making it more suitable for deployment in environments with limited storage capacity. This reduced footprint does not only benefit deployment, but also makes it easier and faster to transfer models across different platforms or networks~\citep{han2016eie}.

    \item Overfitting: Weight pruning can act as a form of regularization, helping to reduce overfitting. By removing less important weights, the model's capacity is reduced, making it less likely to overfit to the training data~\citep{lecun1998gradient}. This can lead to better generalization performance on unseen data.

    \item Performance Trade-off: Despite these benefits, weight pruning must be carefully managed to avoid a significant drop in model performance. If too many weights are pruned, or if the pruning is not conducted carefully, the model can suffer a loss in its capacity to capture complex patterns in the data, resulting in reduced accuracy or other performance metrics~\citep{zhou2018rocket}. 

    \item Fine-tuning Requirement: After pruning, the model often needs to be fine-tuned on the task-specific data to recover any potential loss in performance. This fine-tuning step introduces an additional computational cost and requires careful hyperparameter tuning.
\end{enumerate}

In summary, while weight pruning can lead to increased computational efficiency, reduced model size, and potentially better generalization, it can also introduce a performance trade-off and additional computational costs associated with fine-tuning. It's crucial to carefully consider these factors when implementing weight pruning in Transformer models, ensuring that the pruning strategy aligns well with the specific requirements and constraints of the task and deployment environment.

\section{Related Works}
Natural Language Processing (NLP) has been a major area of research in Artificial Intelligence and Machine Learning
since the early days of computer science~\citep{voorhees1999trec, moldovan2000structure, brill2002analysis, ferrucci2010building, gholami2021zero, gholami2022you, gholami2022create, gholami2022alexa, gholami2022flight, brand2022text, gholami2023generative, gholami2023student}. There are several examples of using pruning to create efficient Transformer models in the literature. In this section we go over a few notable cases.

In addition to being quicker and lightweight than more comprehensive frameworks, more straightforward neural nets are likely to transfer well since it ostensibly retrieves all genuine focus, mainly elements with minimal duplication. Following training, weights with low saliency—those whose removal seems to have a negligible impact upon that loss—can be dropped from extensive systems to shrink their capacity. Approaches that take individual scores into account were considered unorganized, whereas techniques that take attentive units or core networks into account were referred to as organized. There have been numerous suggested unstructured and structured pruning systems, some of which were utilized in Transformers. It is common practice to choose the weights to prune while addressing transformers systems depending on its size or by calculating a significance score using a ﬁrst - order technique. While some techniques significantly reduce the model complexity, the resultant unorganized sparse vectors cannot be used to accelerate inference without dedicated hardware. When pruning pretrained algorithms with challenge tweaking, Sanh et al.~\citep{sanh2020movement} contrast the zeroth- as well as the first pruning techniques. They demonstrate that a straightforward approach for load pruning relying on plain variations is efficient for such a challenge, and it changes by employing a first-order significance score. They use a transformer-based structure to implement this mobility trimming, and empirical results demonstrate that, in high-sparsity regimes, their approach routinely produces significant gains above current approaches.

Sajjat et al.~\citep{sajjad2023effect} discussed various methods for de-layering pre-trained networks and examined the algorithms' overall behavior upon subtasks. To use several pre-trained systems and then a wide range of data sets, researchers ran tests that demonstrated that it is possible to decrease the model complexity and size by nearly 40\% while still retaining close to 98\% of past performance on tasks. The performance of their trimmed algorithms was comparable to that of knowledge-distilled algorithms. Furthermore, their method avoids the need for learning, works with a wide variety of pre-trained versions, especially distilling concepts, and gives users the freedom to choose how much precision and accuracy percent they compromise. Lagunas et al.~\citep{https://doi.org/10.48550/arxiv.2109.04838} have demonstrated that it is possible to obtain tiny trimmed networks equivalent to or superior to distillation systems. This method does not require pre-training and can be utilized while fine-tuning. This solution operates on a wide range of projects or baseline concepts without relying on methods like preprocessing or architectural discovery. Researchers may count on a straightforward and reliable approach for sharing such simulations while maintaining the majority of the actual design correctness when newer and bigger versions are produced at an increased pace. 

Structured pruning, on the other hand, eliminates cohesive score groupings. Recent research demonstrates how certain heads may be eliminated while significantly degrading efficiency, indicating that the majority of heads convey unnecessary features. Regularizing algorithms with structured dropping makes them increasingly resistant to implementing structured pruning at the moment of inference. Fan et al.~\citep{fan2019reducing} concentrate on a scenario in which frameworks were layered, allowing for the pruning of shallower as well as procedures at any appropriate depth. They demonstrate that LayerDrop facilitates and remains stable in ongoing learning on significantly deeper systems inside a range of text creation or pre-training activities. Several scientists have experimented with integrating weight pruning with matrix factorization. Although researchers mix unstructured pruning and Sparsity matrix factorization, Wang et al.~\citep{wang2019structured} propose another all-structural pruning technique in their research that is dependent on adaptable reduced factorization. They assess and evaluate how well this strategy performs on substantial language models. In contrast to existing strategies, such as unstructured amplitude pruning, researchers show that their approach would significantly speed up and condense speeds for extensive systems with no performance degradation. 

Michel et al.~\citep{michel2019sixteen} showed how particular layers might be lowered only to one heading and that numerous units could be deleted into capable transformer systems exhibiting statistical significance efficiency reduction. Furthermore, they demonstrated that the encoder-decoder attention levels in language translation systems were significantly greater dependent on inter-head than in the self-attention stages. Researchers presented proof that the significance of every unit is established during the initial phases. They anticipate that findings might deepen current knowledge of MHA and motivate simulations to use its characteristics and recognition better. Voita et al.~\citep{voita2019analyzing} assess how each attention unit affects the translational efficiency of the Transformer framework. Only a limited selection of units seems crucial for the translated process, as demonstrated by layer-wise significance transmission, which they employ to demonstrate the proportional involvement of units fluctuates. Significant heads perform one or even more decipherable tasks in the framework, such as monitoring certain linguistic relationships and paying attention to nearby words. Researchers offer a novel method for attention head pruning and investigate whether the leftover fewer intelligible units were critical to the architecture efficiency.

\section{Approach}
The approach here emphasizes the elimination of particular weights or connections within the model that have a relatively minimal impact on its performance. By doing so, it is possible to significantly reduce the model's size and enhance computational efficiency, which could also lead to decreased overfitting.

Transformers predominantly consist of self-attention and feed-forward layers. Consequently, these are the primary targets for the application of weight pruning techniques. In this study, we focus on magnitude-based weight pruning and we used the model introduced by~\cite{gholami2023generative} (GPT-Efficio) as the baseline along with bigger GPT-3~\citep{brown2020language} model.

Magnitude-Based Weight Pruning: This technique focuses on the ranking of weights within the Transformer model according to their absolute magnitudes. Those weights exhibiting the smallest magnitudes (i.e., falling beneath a defined threshold) are zeroed or excluded entirely. This approach is premised on the hypothesis that the impact of small-magnitude weights on the model's output is insignificant, and their removal should, therefore, result in a negligible impact on the overall model performance. Subsequent to the pruning process, the model typically undergoes a fine-tuning phase in an attempt to rectify any performance degradation that may have occurred~\citep{molchanov2016pruning}.

In the context of applying weight pruning to Transformer models, it is common practice to prune a minor proportion of weights at any one time, subsequently fine-tuning the model. This iterative pruning process is repeated until the desired degree of sparsity or compression is attained. Such an approach allows the model to incrementally adjust to the removal of various components, thereby mitigating the risk of a significant performance decline.

While weight pruning can assist in reducing the size of the model and the computational demands, it also introduces an additional set of hyperparameters, such as the pruning threshold or rate. These need to be carefully managed to avoid any detrimental impacts~\citep{li2016pruning}. Furthermore, aggressive pruning can lead to a substantial decrease in model performance. As a result, it is of paramount importance to balance the degree of pruning against the performance of the model.

\section{Experiments}

In this section we present the results of our approach in the context of language modeling (i.e. completion tasks) and question answering. 

\subsection{Results}

\begin{table}[!htbp]
\centering
\small
\caption{Performance of pruning technique on completion tasks}\label{pruning-lm}
\begin{tabular}{p{0.23\linewidth} p{0.07\linewidth} p{0.15\linewidth} p{0.15\linewidth} p{0.12\linewidth} p{0.12\linewidth}}
\toprule
\textbf{Model} & \textbf{$n_{params}$} & \textbf{LAMBADA (acc)} & \textbf{LAMBADA (ppl)} & \textbf{StoryCloze (acc)} & \textbf{HellaSwag (acc)} \\ 
\midrule
GPT-3 Zero-Shot & 175B & 76.2 & 3.00 & 83.2 & 78.9   \\ 
GPT-3 One-Shot & 175B & 72.5 & 3.35 & 84.7 & 78.1   \\ 
GPT-3 Few-Shot & 175B & 86.4 & 1.92 & 87.7 & 79.3   \\ 
GPT-Efficio & 950M & 67.1 & 9.2 & 80.5 & 72.6 \\
GPT-Efficio (.3 pruned) & 665M & 61.83 & 10.5 & 74.68 & 64.43 \\
\bottomrule
\end{tabular}
\end{table}

Table~\ref{pruning-lm} demonstrates the GPT-Efficio and pruned GPT-Efficio's performance in comparison with GPT-3. 

\begin{figure}[!htbp]
\centering
\begin{tikzpicture}
\begin{axis}[
    width=\linewidth,
    height=7cm,
    ybar, ymin=0, ymax=100,
    ymin=0,
    xtick=data,
    ylabel=Accuracy (\%),
    xlabel=Model,
    symbolic x coords={GPT-3 Zero-Shot, GPT-3 One-Shot, GPT-3 Few-Shot, GPT-Efficio, GPT-Efficio (.3 pruned)},
    legend style={at={(0.5,-0.15)},
    x tick label style={rotate=12,anchor=east},
    anchor=north,legend columns=-1},
    enlarge x limits=0.2,
    nodes near coords,
    nodes near coords align={vertical},
    every node near coord/.append style={rotate=90, anchor=west},
    legend style={at={(1,1)},
        anchor=south east,legend columns=-1},
    ]
\addplot coordinates {(GPT-3 Zero-Shot, 76.2) (GPT-3 One-Shot, 72.5) (GPT-3 Few-Shot, 86.4) (GPT-Efficio, 67.1) (GPT-Efficio (.3 pruned), 61.83)};
\addplot coordinates {(GPT-3 Zero-Shot, 83.2) (GPT-3 One-Shot, 84.7) (GPT-3 Few-Shot, 87.7) (GPT-Efficio, 80.5) (GPT-Efficio (.3 pruned), 74.68)};
\addplot coordinates {(GPT-3 Zero-Shot, 78.9) (GPT-3 One-Shot, 78.1) (GPT-3 Few-Shot, 79.3) (GPT-Efficio, 72.6) (GPT-Efficio (.3 pruned), 64.43)};
\legend{LAMBADA, StoryCloze, HellaSwag}
\end{axis}
\end{tikzpicture}
\caption{Performance of pruning technique on completion tasks}
\label{fig:pruning-lm}
\end{figure}
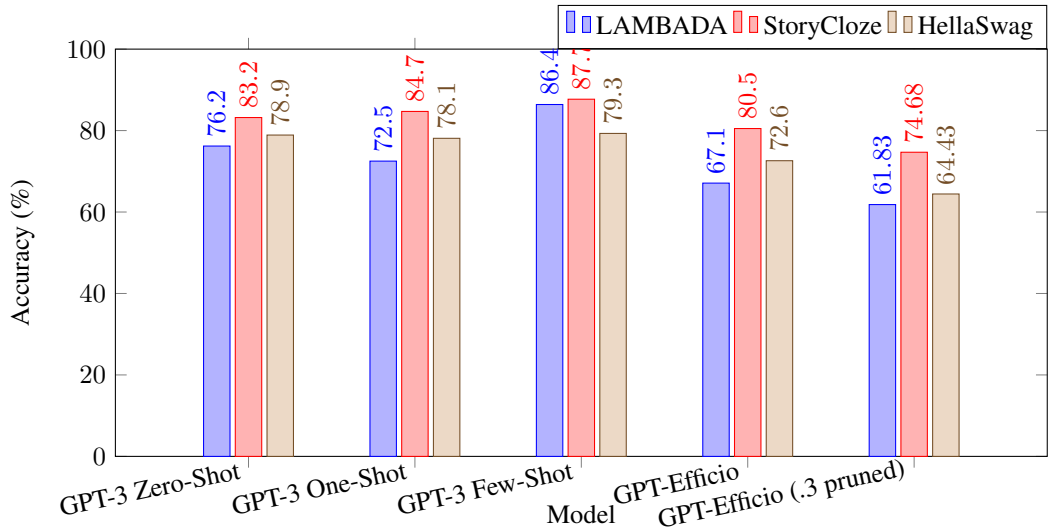

\begin{table}[!htbp]
\centering
\small
\caption{Performance of pruning technique on QA tasks}\label{pruning-qa}
\begin{tabular}{p{0.23\linewidth} p{0.07\linewidth} p{0.15\linewidth} p{0.15\linewidth} p{0.12\linewidth}}
\toprule
\textbf{Model} & \textbf{$n_{params}$} & \textbf{NQ} & \textbf{WebQ} & \textbf{TriviaQA}\\ 
\midrule
GPT-3 Zero-Shot & 175B & 14.6 & 14.4 & 64.3   \\ 
GPT-3 One-Shot & 175B & 23.0 & 25.3 & 68.0   \\ 
GPT-3 Few-Shot & 175B & 29.9 & 41.5 & 71.2   \\ 
GPT-Efficio & 950M & 27.5 & 40.6 & 69.2 \\
GPT-Efficio (.3 pruned) & 665M & 20.35 & 32.72 & 65.43 \\
\bottomrule
\end{tabular}
\end{table}

Table~\ref{pruning-qa} shows the GPT-Efficio and pruned GPT-Efficio's performance in comparison with GPT-3.

\begin{figure}[htbp]
\centering
\begin{tikzpicture}
\begin{axis}[
    width=\linewidth,
    height=7cm,
    ybar,
    ymin=0,
    ylabel={Accuracy},
    xtick=data,
    xticklabels={GPT-3 Zero-Shot, GPT-3 One-Shot, GPT-3 Few-Shot, GPT-Efficio, GPT-Efficio (.3 pruned)},
    nodes near coords,
    nodes near coords align={vertical},
    legend style={at={(0.5,-0.2)},
    anchor=north,legend columns=-1},
    ]
\addplot coordinates {
    (1,14.6) 
    (2,23.0) 
    (3,29.9) 
    (4,27.5) 
    (5,20.35)
    };
\addplot coordinates {
    (1,14.4) 
    (2,25.3) 
    (3,41.5) 
    (4,40.6) 
    (5,32.72)
    };
\addplot coordinates {
    (1,64.3) 
    (2,68.0) 
    (3,71.2) 
    (4,69.2) 
    (5,65.43)
    };
\legend{NQ, WebQ, TriviaQA}
\end{axis}
\end{tikzpicture}
\caption{Performance of pruning technique on QA tasks}
\label{fig:pruning-qa}
\end{figure}
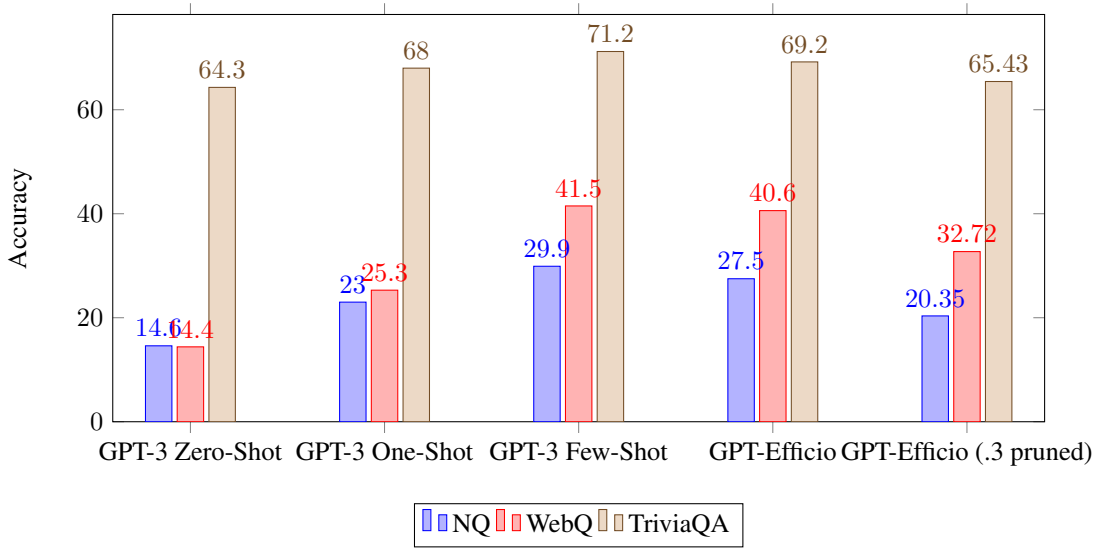


\section{Analysis}
The hyperparameters related to weight pruning in large language models can significantly influence the outcome of the pruning process and ultimately the performance of the pruned model. Here are some of the key hyperparameters:

\begin{enumerate}
    \item Pruning Threshold: This is a hyperparameter that determines the cutoff point for weights to be pruned. For magnitude-based pruning, weights whose absolute values are below this threshold will be pruned. The choice of threshold will directly affect the degree of pruning and consequently the size and speed of the model. However, a too aggressive threshold could remove important weights and degrade model performance~\citep{han2015deep}.

    \item Pruning Rate: This hyperparameter defines the proportion of weights to be pruned in each pruning iteration. For example, a pruning rate of 0.2 would mean that 20\% of the weights are pruned in each iteration. The pruning rate affects the speed at which pruning is performed and the final level of sparsity in the model~\citep{molchanov2016pruning}. Similar to the pruning threshold, a high pruning rate may lead to faster model size reduction but could also result in significant performance loss if it's too high.

    \item Pruning Schedule: This parameter dictates how pruning is carried out over the course of training or fine-tuning. It could be constant, linear, or exponential, among others. For instance, in a linear pruning schedule, an increasing proportion of weights would be pruned as training progresses. The pruning schedule can influence how the model adapts to the pruning process and can have a substantial impact on the final model performance.

    \item Regularization Parameter: In regularization-based pruning methods, the regularization parameter (usually denoted as lambda in L1 regularization) controls the degree of penalty applied to the weights. A larger regularization parameter encourages sparser weight matrices but might also lead to performance degradation if set too high.

    \item Retraining or Fine-tuning epochs: After each pruning iteration, the model is typically retrained or fine-tuned for a certain number of epochs. The number of these epochs is also a hyperparameter that can influence how well the model recovers from the pruning process. Too few epochs might not allow the model to adequately adapt to the pruned structure, while too many epochs can be computationally expensive~\citep{li2016pruning}.

    \item Initial Sparsity and Target Sparsity: These hyperparameters are typically used in iterative pruning methods. Initial sparsity is the proportion of weights to prune at the beginning, and target sparsity is the final desired proportion of weights to prune. The values of these hyperparameters can directly affect the efficiency and performance of the pruned model.
\end{enumerate}

All these hyperparameters must be carefully tuned to strike a balance between model size, computational efficiency, and model performance. It's crucial to conduct systematic experiments or leverage hyperparameter optimization techniques to find the optimal configuration for these parameters. The optimal set of hyperparameters can be task-specific and might vary depending on the specific requirements and constraints of the task and deployment environment.

In this section, we focus on the pruning rate hyperparameter. The pruning rate, one of the primary hyperparameters in weight pruning techniques, significantly influences the performance of a language model such as a Transformer.

The pruning rate essentially determines the proportion of weights to be pruned in each pruning iteration. For example, if you set a pruning rate of 0.1, it means that in each pruning step, 10\% of the weights (usually the ones with the smallest magnitudes) will be eliminated.

The effect of the pruning rate on the performance of the language model can be viewed from several angles:

\begin{enumerate}
    \item Model Performance: Higher pruning rates might lead to a more substantial reduction in the model size, which is beneficial for computational efficiency~\citep{neklyudov2017structured}. However, pruning a large number of weights at once can potentially result in the loss of important information that the model needs to make accurate predictions. As such, if the pruning rate is set too high, it can substantially drop model performance. Conversely, a lower pruning rate can preserve more information, reducing the risk of performance degradation but resulting in a less compact model.

    \item Fine-tuning Requirements: The higher the pruning rate, the more dramatic the change to the model structure, meaning the model may require more extensive fine-tuning to recover from the pruning process. Therefore, a high pruning rate may increase the computational cost of the fine-tuning process and the complexity of finding the right fine-tuning hyperparameters.

    \item Overfitting: In certain scenarios, a higher pruning rate can act as a form of regularization. By reducing the number of parameters, the model's capacity to memorize the training data is reduced, potentially mitigating overfitting and improving the model's ability to generalize to unseen data. However, it's important to note that this effect has its limits – pruning too aggressively can degrade model performance by removing too much capacity.

    \item Convergence Speed: Typically, with a lower pruning rate, the model's convergence speed during training or fine-tuning can be maintained since the model changes less drastically between iterations. On the other hand, a high pruning rate can potentially disrupt the learning process and slow down convergence, as the model needs to adapt to more substantial changes at each step.
\end{enumerate}

In conclusion, the choice of pruning rate involves a trade-off between model performance, computational efficiency, fine-tuning requirements, and the potential for regularization. Careful experimentation and tuning are required to select an optimal pruning rate that suits the specific demands of your task and environment.

\begin{table}[!htbp]
\centering
\small
\caption{Analysis of the effects of hyperparameter pruning rate on completion tasks}\label{pruning-anal-lm}
\begin{tabular}{p{0.18\linewidth} p{0.06\linewidth} p{0.06\linewidth} p{0.15\linewidth} p{0.15\linewidth} p{0.12\linewidth} p{0.12\linewidth}}
\toprule
\textbf{Model} & \textbf{pr\%} & \textbf{$n_{params}$} & \textbf{LAMBADA (acc)} & \textbf{LAMBADA (ppl)} & \textbf{StoryCloze (acc)} & \textbf{HellaSwag (acc)} \\ 
\midrule
GPT-3 Zero-Shot & - & 175B & 76.2 & 3.00 & 83.2 & 78.9   \\ 
GPT-3 One-Shot & - & 175B & 72.5 & 3.35 & 84.7 & 78.1   \\ 
GPT-3 Few-Shot & - & 175B & 86.4 & 1.92 & 87.7 & 79.3   \\ 
GPT-Efficio & - & 950M & 67.1 & 9.2 & 80.5 & 72.6 \\
GPT-Efficio & .1 & 855M & 67.05 & 9.4 & 80.21 & 72.16 \\
GPT-Efficio & .3 & 665M & 61.83 & 10.5 & 74.68 & 64.43 \\
GPT-Efficio & .5 & 475M & 55.38 & 12.64 & 65.50 & 52.68 \\
\bottomrule
\end{tabular}
\end{table}

Table~\ref{pruning-anal-lm} demonstrates the GPT-Efficio and pruned GPT-Efficio's performance in comparison with GPT-3. 

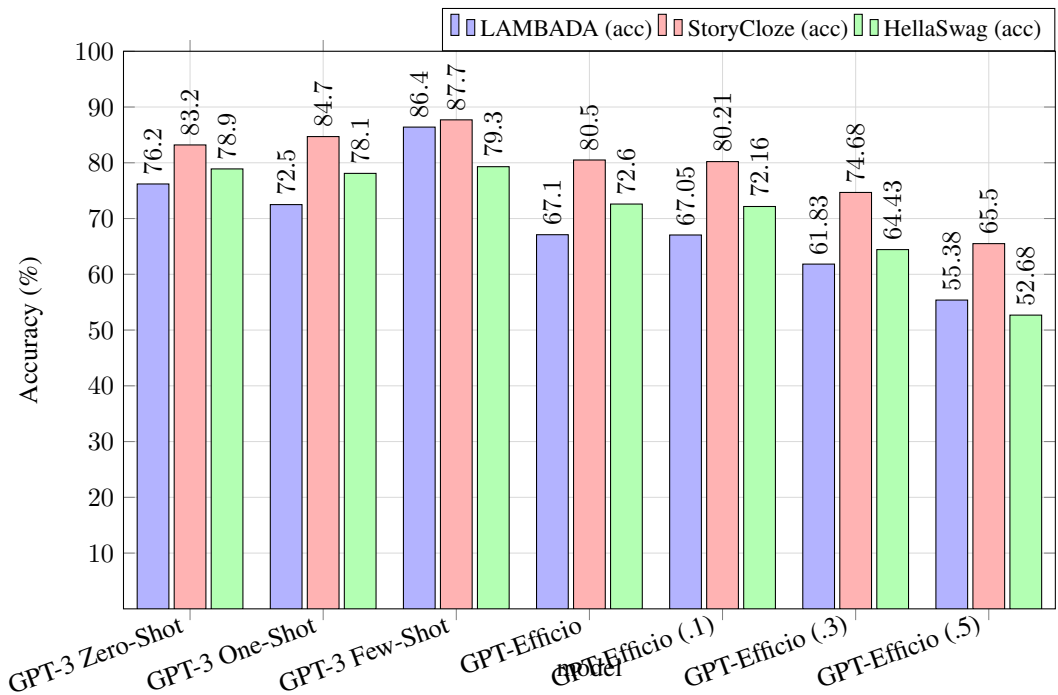
\begin{figure}[!htbp]
\centering
\begin{tikzpicture}
\begin{axis}[
    ybar, ymin=0, ymax=100,
    bar width=12pt,
    width=\linewidth,
    height=9cm,
    xlabel={model},
    ylabel={Accuracy (\%)},
    xmin=0.5, xmax=7.5,
    ymin=0, ymax=100,
    xtick={1, 2, 3, 4, 5, 6,7},
    xticklabels={GPT-3 Zero-Shot, GPT-3 One-Shot, GPT-3 Few-Shot, GPT-Efficio , GPT-Efficio (.1), GPT-Efficio (.3), GPT-Efficio (.5)},
    ytick={10, 20, 30, 40, 50, 60, 70, 80, 90, 100},
    legend style={at={(0.5,-0.15)},anchor=north,legend columns=-1},
    grid=both,
    grid style={line width=0.2pt, draw=gray!30},
    xmajorgrids=true,
    ymajorgrids=true,
    xticklabel style={rotate=20, anchor=east},
    nodes near coords,
    nodes near coords align={vertical},
    every node near coord/.append style={rotate=90, anchor=west},
    legend style={font=\footnotesize},
    legend style={at={(1,1)},
        anchor=south east,legend columns=-1}
]
\addplot[fill=blue!30] coordinates {
    (1, 76.2)
    (2, 72.5)
    (3, 86.4)
    (4, 67.1)
    (5, 67.05)
    (6, 61.83)
    (7, 55.38)
};
\addplot[fill=red!30] coordinates {
    (1, 83.2)
    (2, 84.7)
    (3, 87.7)
    (4, 80.5)
    (5, 80.21)
    (6, 74.68)
    (7, 65.50)
};
\addplot[fill=green!30] coordinates {
    (1, 78.9)
    (2, 78.1)
    (3, 79.3)
    (4, 72.6)
    (5, 72.16)
    (6, 64.43)
    (7, 52.68)
};
\legend{LAMBADA (acc), StoryCloze (acc), HellaSwag (acc)}
\end{axis}
\end{tikzpicture}
\caption{Analysis of the effects of the pruning approach on completion tasks.}
\label{fig:param-anal-qa}
\end{figure}

\begin{table}[!htbp]
\centering
\small
\caption{Analysis of the effects of hyperparameter pruning rate on QA tasks}\label{pruning-anal-qa}
\begin{tabular}{p{0.18\linewidth} p{0.06\linewidth} p{0.06\linewidth} p{0.15\linewidth} p{0.15\linewidth} p{0.12\linewidth}}
\toprule
\textbf{Model} & \textbf{pr\%} & \textbf{$n_{params}$} & \textbf{NQ} & \textbf{WebQ} & \textbf{TriviaQA}\\ 
\midrule
GPT-3 Zero-Shot & - & 175B & 14.6 & 14.4 & 64.3   \\ 
GPT-3 One-Shot & - & 175B & 23.0 & 25.3 & 68.0   \\ 
GPT-3 Few-Shot & - & 175B & 29.9 & 41.5 & 71.2   \\ 
GPT-Efficio & - & 950M & 27.5 & 40.6 & 69.2 \\
GPT-Efficio & .1 & 855M & 26.35 & 39.56 & 68.76 \\
GPT-Efficio & .3 & 665M & 22.39 & 37.43 & 65.91 \\
GPT-Efficio & .5 & 475M & 19.83 & 32.76 & 60.84 \\
\bottomrule
\end{tabular}
\end{table}

Table~\ref{pruning-anal-qa} shows the GPT-Efficio and pruned GPT-Efficio's performance in comparison with GPT-3.

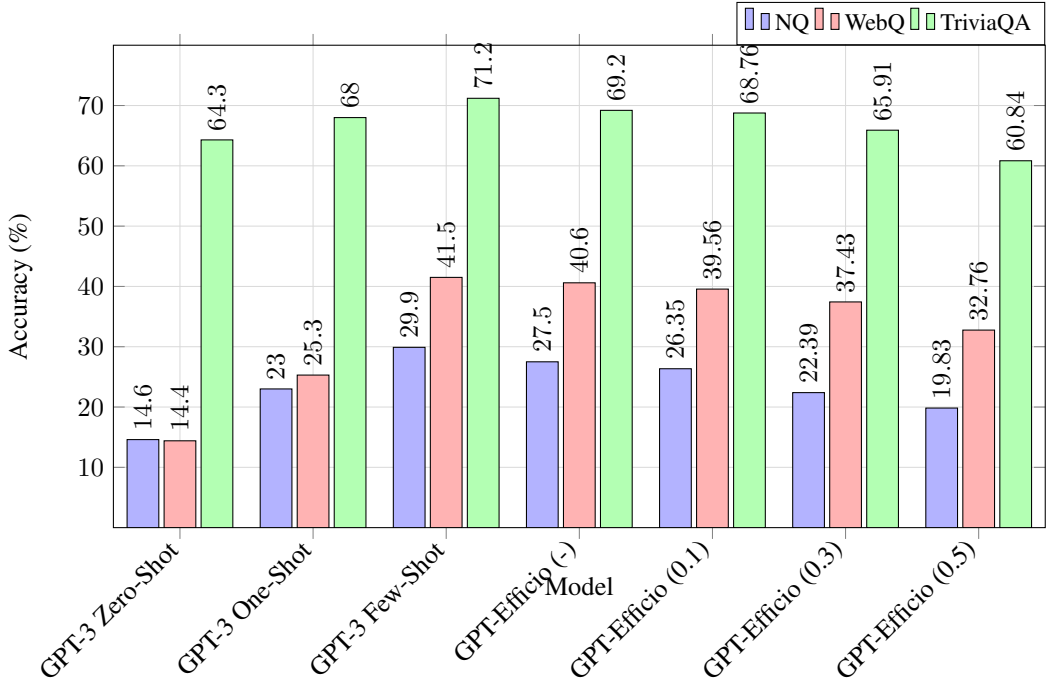
\begin{figure}[!htbp]
\centering
\begin{tikzpicture}
\begin{axis}[
    ybar, ymin=0, ymax=100,
    bar width=12pt,
    width=\linewidth,
    height=8cm,
    xlabel={Model},
    ylabel={Accuracy (\%)},
    xmin=0.5, xmax=7.5,
    ymin=0, ymax=80,
    xtick={1, 2, 3, 4, 5, 6, 7},
    xticklabels={GPT-3 Zero-Shot, GPT-3 One-Shot, GPT-3 Few-Shot, GPT-Efficio (-), GPT-Efficio (0.1), GPT-Efficio (0.3), GPT-Efficio (0.5)},
    ytick={10, 20, 30, 40, 50, 60, 70},
    legend style={at={(0.5,-0.15)},anchor=north,legend columns=-1},
    grid=both,
    grid style={line width=0.2pt, draw=gray!30},
    xmajorgrids=true,
    ymajorgrids=true,
    xticklabel style={rotate=45, anchor=east},
    nodes near coords,
    nodes near coords align={vertical},
    every node near coord/.append style={rotate=90, anchor=west},
    legend style={font=\footnotesize},
    legend style={at={(1,1)},
        anchor=south east,legend columns=-1}
]
\addplot[fill=blue!30] coordinates {
    (1, 14.6)
    (2, 23.0)
    (3, 29.9)
    (4, 27.5)
    (5, 26.35)
    (6, 22.39)
    (7, 19.83)
};
\addplot[fill=red!30] coordinates {
    (1, 14.4)
    (2, 25.3)
    (3, 41.5)
    (4, 40.6)
    (5, 39.56)
    (6, 37.43)
    (7, 32.76)
};
\addplot[fill=green!30] coordinates {
    (1, 64.3)
    (2, 68.0)
    (3, 71.2)
    (4, 69.2)
    (5, 68.76)
    (6, 65.91)
    (7, 60.84)
};
\legend{NQ, WebQ, TriviaQA}
\end{axis}
\end{tikzpicture}
\caption{Analysis of the effects of hyperparameter pruning rate on QA tasks}
\label{fig:pruning-anal-qa}
\end{figure}

\section{Limitations}
While weight pruning in Transformer models can lead to significant benefits in terms of model size reduction and computational efficiency, it does come with a set of limitations:

\begin{enumerate}
    \item Performance Trade-off: One of the major limitations of pruning is the trade-off between model size and performance. Although pruning can reduce the number of parameters in the model, it also reduces the model's capacity, which can lead to a drop in performance. If pruning is too aggressive, it can result in substantial information loss and significantly degrade the model's accuracy.

    \item Complexity in Hyperparameter Tuning: Pruning introduces additional hyperparameters such as the pruning rate, threshold, and schedule, which need to be tuned carefully. Incorrect setting of these hyperparameters can lead to ineffective pruning or significant performance degradation. Determining the optimal values for these hyperparameters can be a complex and resource-intensive task that requires systematic experimentation or sophisticated optimization strategies.

    \item Post-Pruning Fine-Tuning: After pruning, the model usually needs to be fine-tuned to recover any performance loss caused by the pruning process. This fine-tuning step introduces an additional computational cost and requires careful hyperparameter tuning.

    \item Sparsity-Induced Inefficiency: Standard deep learning frameworks and hardware accelerators are optimized for dense matrix operations. While pruning introduces sparsity in the weight matrices, which theoretically should speed up computation, these sparse operations are not as efficiently supported, leading to a potential under-utilization of computational resources. This issue is being addressed by ongoing research and development in hardware and software for sparse computations.

    \item Irregular Sparsity Patterns: Pruning often leads to irregular sparsity patterns that are hard to exploit for computational efficiency. In contrast, structured pruning methods (which prune entire neurons, layers, or channels) create regular sparsity patterns but are more restrictive and might result in higher accuracy loss.

    \item Dependence on Initial Training: The effectiveness of pruning can heavily depend on the initial training of the model. If the model is not well trained before pruning, the pruning process may remove important weights, leading to substantial performance degradation.
\end{enumerate}

While pruning can provide significant advantages, these limitations indicate that it's not a silver bullet. The benefits of pruning must be carefully weighed against its potential downsides, and the process must be carefully managed to ensure optimal performance.

\section{Future Work}
While weight pruning has demonstrated effectiveness in reducing the size and improving the efficiency of Transformer models, there's much scope for future work to further enhance this approach and overcome its limitations. Here are a few suggestions:

\begin{itemize}
    \item Advanced Pruning Strategies: Develop more advanced and nuanced pruning strategies that can better balance model size reduction and performance. This could involve new methods for deciding which weights to prune, beyond just considering weight magnitude.

    \item Pruning with Context: Investigate context-aware pruning strategies that take into account the importance of weights in relation to specific inputs or tasks. For example, certain weights may be more critical for specific types of inputs or tasks and should be preserved.

    \item Regularization Techniques: Explore the integration of pruning with other regularization techniques to limit the performance drop after pruning. This could help enhance the model's robustness and generalization capability while also reducing its size.

    \item Improved Fine-tuning Strategies: Develop more effective fine-tuning strategies for post-pruning model recovery. This could involve adaptive learning rates or other methods to accelerate the fine-tuning process and reduce its computational cost.

    \item Sparse Training: Investigate methods for sparse training, which involves introducing sparsity during the initial training process rather than post hoc. Sparse training could lead to models that are naturally smaller and more efficient, and potentially avoid some of the performance degradation associated with pruning.

    \item Hardware and Software for Sparse Computations: Contribute to the development of hardware and software that can more effectively handle the sparse computations resulting from pruning. This would help to realize the computational benefits of pruning and make pruned models more efficient to run on various devices.

    \item Structured Pruning: Expand on structured pruning methods, which remove larger, coherent parts of the model, such as entire neurons or layers. While these methods can lead to more significant size reductions and are more amenable to existing hardware accelerators, they often result in a greater loss in accuracy and require further research.

    \item Pruning with Knowledge Distillation: Combining pruning with knowledge distillation, where the knowledge from a larger model (teacher) is transferred to a smaller model (student), could be a promising avenue for maintaining performance while reducing model size.

    \item Understanding Pruning: Improve our theoretical understanding of why and how pruning works. Despite its empirical success, the underlying theory of why pruning does not drastically harm model performance and sometimes even improves it is not fully understood.
\end{itemize}

These suggestions highlight the potential to further improve pruning techniques and better integrate them into the deep learning pipeline, increasing the efficiency and practicality of Transformer models.

\section{Conclusion}
The allure of Transformer models lies in their unmatched prowess in modeling intricate patterns and relationships in data, especially within the realm of natural language processing. Nevertheless, their expansive parameterization has given rise to palpable challenges in computational efficiency, scalability, and ecological sustainability. Through the lens of this research, we have navigated the intricate landscape of weight pruning as a potential panacea to these challenges. Our systematic exploration affirms that with an astute pruning strategy, it is indeed feasible to achieve a harmonious balance between model size reduction and performance retention. Notably, post-pruning fine-tuning emerged as a salient phase, occasionally enhancing the generalization capabilities of pruned models beyond their dense counterparts.

As the deep learning community continues its relentless pursuit of excellence, considerations of model efficiency and environmental stewardship must ascend to the forefront of our collective consciousness. Weight pruning, as showcased in this paper, offers a promising avenue in this regard. However, it is not a terminal solution. It beckons further refinement and needs to be integrated with other optimization techniques for holistic model improvement. As we stand on the cusp of this research frontier, the call is clear: the future of deep learning not only demands models that think more profoundly but also those that think more efficiently.

\bibliographystyle{plainnat}
\bibliography{main}

\end{document}